\documentclass[11pt,a4paper]{article}
\usepackage[hyperref]{naaclhlt2018}
\usepackage{times}
\usepackage{latexsym}
\usepackage[inline]{enumitem}
\usepackage{amsfonts, amsmath}
\usepackage{graphicx}  %Required
\usepackage{multirow}
\usepackage{boldline}
\usepackage{cleveref}

\usepackage{url}

\aclfinalcopy % Uncomment this line for the final submission
\title{Efficient Graph-based Word Sense Induction \\ by Distributional Inclusion Vector Embeddings}

\author{Haw-Shiuan Chang$^1$, Amol Agrawal$^1$, Ananya Ganesh$^1$, \\
\textbf{Anirudha Desai$^1$, Vinayak Mathur$^1$, Alfred Hough$^2$, Andrew McCallum$^1$}\\
$^1$CICS, University of Massachusetts, 140 Governors Dr., Amherst, MA 01003 \\
$^2$Lexalytics, 320 Congress St, Boston, MA 02210\\
\\
\texttt{\{hschang,amolagrawal,aganesh\}@cs.umass.edu,}\\
\texttt{\{anirudhadesa,vinayak,mccallum\}@cs.umass.edu} \\
\texttt{al.hough@lexalytics.com}
}

\date{}

\newcommand{\smallt}[1]{{\normalsize{\sf #1}}}
\setlist[itemize]{leftmargin=*}

\crefformat{footnote}{#2\footnotemark[#1]#3}

\begin{document}
\maketitle
\begin{abstract}
%Polysemy often causes trouble in the natural language processing (NLP) systems. 
Word sense induction (WSI), which addresses polysemy by unsupervised discovery of multiple word senses, resolves ambiguities for downstream NLP tasks and also makes word representations more interpretable. This paper proposes an accurate and efficient graph-based method for WSI that builds a global non-negative vector embedding basis (which are interpretable like topics) and clusters the basis indexes in the ego network of each polysemous word. By adopting {\it distributional inclusion vector embeddings} as our basis formation model, we avoid the expensive step of nearest neighbor search that plagues other graph-based methods without sacrificing the quality of sense clusters.  Experiments on three datasets show that our proposed method produces similar or better sense clusters and embeddings compared with previous state-of-the-art methods while being significantly more efficient.
%using distributional
\end{abstract}

% Word sense induction (WSI), which addresses polysemy by unsupervised discovery multiple word senses, not only resolves ambiguities for downstream NLP tasks, but also facilitates our understanding of a corpus but also make word representation more intuitive, interpretable and potentially more generalizable. This paper proposes an efficient graph-based method for WSI that builds a global topic model and clusters topics in the ego network of each target word. By adopting distributional inclusion vector embedding as our topic model, we can avoid the expensive step of nearest neighbor search in other graph-based methods without sacrificing the quality of sense clusters. The experiments on three datasets show that proposed method performs similar or better compared with a state-of-the-art graph-based WSI approach.

\section{Introduction}
%\add[HS]{new text} 
%\remove[HS]{existing text}
%\change[HS]{existing text}{new text}
%\annote[HS]{existing text}{note}
%\note[HS]{note}
\label{sec:intro}
Word sense induction (WSI) is a challenging task of natural language processing whose goal is to categorize and identify multiple senses of polysemous words from raw text without the help of predefined sense inventory like WordNet~\citep{miller1995wordnet}. The problem is sometimes also called unsupervised word sense disambiguation~\citep{AgirreMLS06,PelevinaABP16}. 

An effective WSI has wide applications. For example, we can 
compare different induced senses in different documents to detect novel senses over time~\citep{LauCMNB12,MitraMRBMG14} or analyze sense difference in multiple corpora~\citep{MathewMSMG17}. WSI could also be used to group and diversify the documents retrieved from search engine~\citep{navigli2010inducing,MarcoN13}. After identifying senses, we can train an embedding for each sense of a word. \citet{LiJ15} demonstrate that this multi-prototype word embedding is useful in several downstream applications including part-of-speech (POS) tagging, relation extraction, and sentence relatedness tasks. \citet{sumanth2015much} also show that word sense disambiguation could be successfully applied to sentiment analysis.

%Novel sense by topic modeling\citep{LauCMNB12}

%graph based novel sense\citep{LauCMNB12,MitraMRBMG14}
%identifying corpus-specific sense differences\citep{MathewMSMG17}

%graph-based
%Help search\citep{navigli2010inducing,MarcoN13}

%Embedding applications
%\citep{LiJ15}

Since word sense induction (WSI) methods are unsupervised, the senses are typically derived from the results of different clustering techniques. Like most of the clustering problems, it is usually challenging to predetermine the number of clusters/senses each word should have. In fact, for many words, the ``correct" number of senses is not unique. Setting the number of clusters differently can capture different resolutions of senses. For instance, \smallt{race} in the \smallt{car} context could share the same sense with the \smallt{race} in the \smallt{game} context because they all mean contest, but the \smallt{race} in the \smallt{car} context actually refers to the specific contest of speed. Therefore, they can also be separated into two different senses, depending on the level of granularity we would like to model.

For graph-based clustering methods, it is easy and natural to model the multiple resolutions of senses in a consistent way by hierarchical clustering and defer the difficult problem of choosing the number of clusters to the end. This makes it easier to incorporate other information, such as users' resolution preference on each hierarchical sense tree. The flexibility is one of the reasons why graph-based methods are widely studied and applied to many downstream applications~\citep{MitraMRBMG14,MathewMSMG17,navigli2010inducing,MarcoN13}.

Nevertheless, graph-based WSI methods usually require a substantial amount of computational resources. For example, \citet{PelevinaABP16} build the graph by finding the nearest neighbors of the target word in the word embedding space (i.e., ego network). Thus, constructing ego networks for all the words takes at least $O(|V|^2)$ time, where $|V|$ is the size of the vocabulary, unless some approximation is made (e.g., approximate nearest neighbor search such as k-d tree).\footnote{\citet{PelevinaABP16} also suggest that JoBimText is an efficient alternative to estimating word similarity, but the method still needs time to run a dependent parser and not every domain has an efficient and high-quality parser.} Next, if our goals include finding less common senses, the method needs to construct a large graph by including more nearest neighbors. For each target word, computing the pairwise distances between nodes in the large graph is also computationally intensive. 
%One common disadvantage of 
%is that the methods

To overcome the limitations and make graph-based WSI more practical, we propose a novel WSI algorithm that first groups words into a set of basis indexes (i.e., a set of topics) efficiently and then, constructs the graph where each node corresponds to a basis index (i.e., a topic) instead of a word. The motivation behind the approach is that different senses of a word usually appear in different topics. For example, \smallt{food} and \smallt{technology} will be at least two distinct topics in most of the topic models, so we can find senses by clustering corresponding basis indexes safely when the target word is \smallt{apple}. If one word could have distinct senses in one topic, humans will constantly face difficult word sense disambiguation tasks while reading a document.

%do not group words related to food and words related to technology into the same topic, so .

Although the main idea is simple,  improving the efficiency significantly without sacrificing the quality is difficult. One of the challenges is that similarity between two basis indexes changes given different target words. For example, a \smallt{country} topic should be clustered together with a \smallt{city} topic if the target word is \smallt{place}. However, if the query word is \smallt{bank}, it makes more sense to group the \smallt{country} topic with the \smallt{money} topic into one sense so that the bank mention in \smallt{Bank of America} will belong to the sense. This means we want to focus on the geographical meaning of \smallt{country} when the target word is more about geography, while focus on the economic meaning of \smallt{country} when the target word is more about economics.

In order to tackle the issue, we adopt a recently proposed approach called distributional inclusion vector embedding (DIVE)~\citep{chang2017unsupervised}. DIVE compresses the sparse bag-of-words while preserving the co-occurrence frequency order, so DIVE is able to model not only the possibility of observing one target word in a topic as typical topic models but also the possibility of observing one topic of a sentence containing a target word mention. This allows us to efficiently identify the topics relevant to each target word, and only focus on an aspect of each of these topics composed of the words relevant to both the topic and the target word.
%topics relevant words for these topics, and only focus on relevant words %, and judge the 

%so that each word has higher embedding values on a basis index when the word occurs more frequently in the corresponding topic. This allows The latter 
% This allows us to efficiently identify the relevant words which characterize the query word, and estimate the target-dependent basis similarity based on those words.
%DIVE not only creates a basis
%topic model but also preserves the frequency of target word in each topic by compressing sparse bag of word. 

Experiments show that our method performs similarly compared with \citet{PelevinaABP16}, a state-of-the-art graph-based WSI method, without the need of expensive nearest neighbor search. Our method is even better for the words without a dominating sense.

% as our topic model
%access the general words related to the query word and .

%However, clustering on the global features might group topics together based on the co-occurrence of words which are unrelated to the query words and we want to make the similarity dependent on the query word. 

%like we did in the word sense disambiguation experiment (Table~\ref{tb:WSD}). % in the paper. 

%there are several difficulties that need to be overcome. 
% This is because 

%One of main reasons that existing graph-based method works well is that we only cluster the words related to the target word.

%Finally, when performing the graph clustering, the connection between each node and the potential contexts of each mention
%Out method cluster topic-> can quickly retrieve relevant mentions or documents on the fly

%how many nearest neighbors 
%The nodes in the graph could be roughly categorized into 
%, and 

%This means the correct number of clusters is not unique, and the methods, which fixes the cluster numbers, need to re-train the embedding many times to capture such granularity.

%In the word sense induction tasks, 

%Many applications choose to use graph-based methods

% (like the results in Table~\ref{tb:WSD}).

%graph based clustering words similar to the target word\citep{lin1998information,pantel2002discovering,DorowW03,biemann2006chinese,hope2013maxmax,PelevinaABP16}

\section{Related Work}
%There are different ways to constructing the graph and clustering its nodes. 
WSI methods can be roughly divided into two categories~\citep{PelevinaABP16}: clustering words similar to the target/query word or clustering mentions of the target word. We address their general limitations below.
%\footnote{Few methods might develop some heuristics to alleviate some of the issues. However, comparing the pros and cons of these detailed designs is out of the scope of this study.}

%Different graph-based methods may have different pros and cons, but 
\subsection{Clustering Related Words}
Graph-based clustering for WSI has a long history and many different variations~\citep{lin1998information,pantel2002discovering,DorowW03,Veronis04,AgirreMLS06,biemann2006chinese,navigli2010inducing,hope2013maxmax,MarcoN13,MitraMRBMG14,PelevinaABP16}. In general, the method is to first retrieve words similar or related to each target word as nodes, measure the similarity/relatedness between the words to form an ego graph/network, and either group the nodes by graph clustering or find hubs or representative nodes in the graph using HyperLex~\citep{Veronis04} or PageRank~\citep{AgirreMLS06}. 

As we mentioned in the introduction section, building word similarity graph and performing graph clustering is usually computationally expensive unless relying on information other than co-occurrence statistics such as word snippets from a search engine~\citep{navigli2010inducing,MarcoN13} or existing high-quality dependency parse~\citep{MitraMRBMG14,PelevinaABP16}. Depending on the downstream applications and word similarity estimation algorithms available at the time of each work, the methods strive for the balance between efficiency and quality in different ways.

Most of the WSI methods that cluster words use graph-based algorithms. One notable exception is \citet{LauCMNB12}. For each target word, they build a topic model, latent Dirichlet allocation or its extension, on the contexts of all mentions of target words. Although computing pairwise similarity is not required here, the approach is still computationally expensive because there might be tens of thousands of mentions of a target word in the corpus and the approach needs to train $V$ different topic models instead of globally modeling topics once like our method. 
%just training a global topic model like our method. 

In addition to the scalability concerns, we do not know how many mentions of a target word are semantically closest to each of its most related words (i.e., node in its ego-network). The loss of connection makes balance the cluster size during the clustering difficult. Furthermore, it might be common that when users would like to adopt fine-grained senses in the hierarchical clustering tree but realize that there is no mention in the corpus that would be categorized into some sense clusters.

%causes difficulty if we would like to balance the cluster size during the clustering

%graph based clustering words similar to the target word

%graph based
%Help search\citep{navigli2010inducing,MarcoN13}

%finding hubs
%HyperLex\citep{Veronis04}
%PageRank\citep{AgirreMLS06}
%PageRank\citep{AgirreMLS06,AgirreS09}

%In addition to the time
%building network takes time

\subsection{Clustering Mentions}
In addition to clustering words similar/related to the target word, we can also cluster every mention based on its context words, which co-occur in a small window. Although this way saves the time of finding similar words, the samples need to be clustered drastically increase because each target word could have tens of thousands of mentions in the corpus of interest. This makes bottom-up hierarchical clustering or global optimization such as spectral clustering~\citep{stella2003multiclass} become infeasible. Without hierarchical sense clustering, it is hard to inject other sources of information such as user intervention or prior knowledge to determine the number of clusters.

To efficiently cluster many samples, \citet{schutze1992dimensions} sub-samples the context of mentions; \citet{mu2016geometry} run principle component analysis (PCA) to compress the contexts of each target word before clustering; other approaches adopt iteratively local search algorithms after random initialization such as expectation maximization (EM)~\citep{ReisingerM10,NeelakantanSPM14,tian2014probabilistic,PinaJ15,LiJ15,BartunovKOV16} or gradient descent~\citep{athiwaratkun2017multimodal}. Although the random initialization and local search methods could be very efficient, the methods might suffer from bad local minimums. Moreover, the users need to specify the number of senses or a global hyper-parameter which controls the level of granularity at the beginning and hope that it will output the sense models with desired resolution after training finishes. The lack of a way to browsing different sense resolution limits the application of the type of WSI methods.
%\citep{schutze1992dimensions,ReisingerM10,NeelakantanSPM14,tian2014probabilistic,PinaJ15,LiJ15,BartunovKOV16,mu2016geometry,athiwaratkun2017multimodal}

%Speed
%Some of the method do different things when target words are %different. Slow like graph-based method
%clustering context of each word separately take lots of time
%\citep{mu2016geometry}

%number of samples is larger. Hard to do bottom up hierarchical clustering
%but do not have hierarchical senses. 

%especially when optimize embedding directly, need to retrain the embedding if number of senses changes
%EM based embedding
%MSSG\citep{NeelakantanSPM14,tian2014probabilistic,PinaJ15,LiJ15,BartunovKOV16}
%Multiple Gaussian embedding\citep{athiwaratkun2017multimodal}

%Speed -> Low Quality
%if want to top down, hard to optimize globally
%random initialize\citep{ReisingerM10,NeelakantanSPM14,tian2014probabilistic,PinaJ15,LiJ15} or random subsample~\citep{schutze1992dimensions}.

%In the word sense disambiguation tasks, it is usually challenging to determine how many senses/clusters each word should have. Many existing approaches fix the number of senses before training the embedding~\citep{tian2014probabilistic,athiwaratkun2017multimodal}. \citet{NeelakantanSPM14} make the number of clusters approximately proportional to the diversity of the context, but the assumption does not always hold. Furthermore, the training process cannot capture different granularity of senses. 

%soft clustering\citep{LauCMNB12,mu2016geometry}

\section{Method}

\begin{figure*}[t!]
\centering
\includegraphics[width=0.8\linewidth]{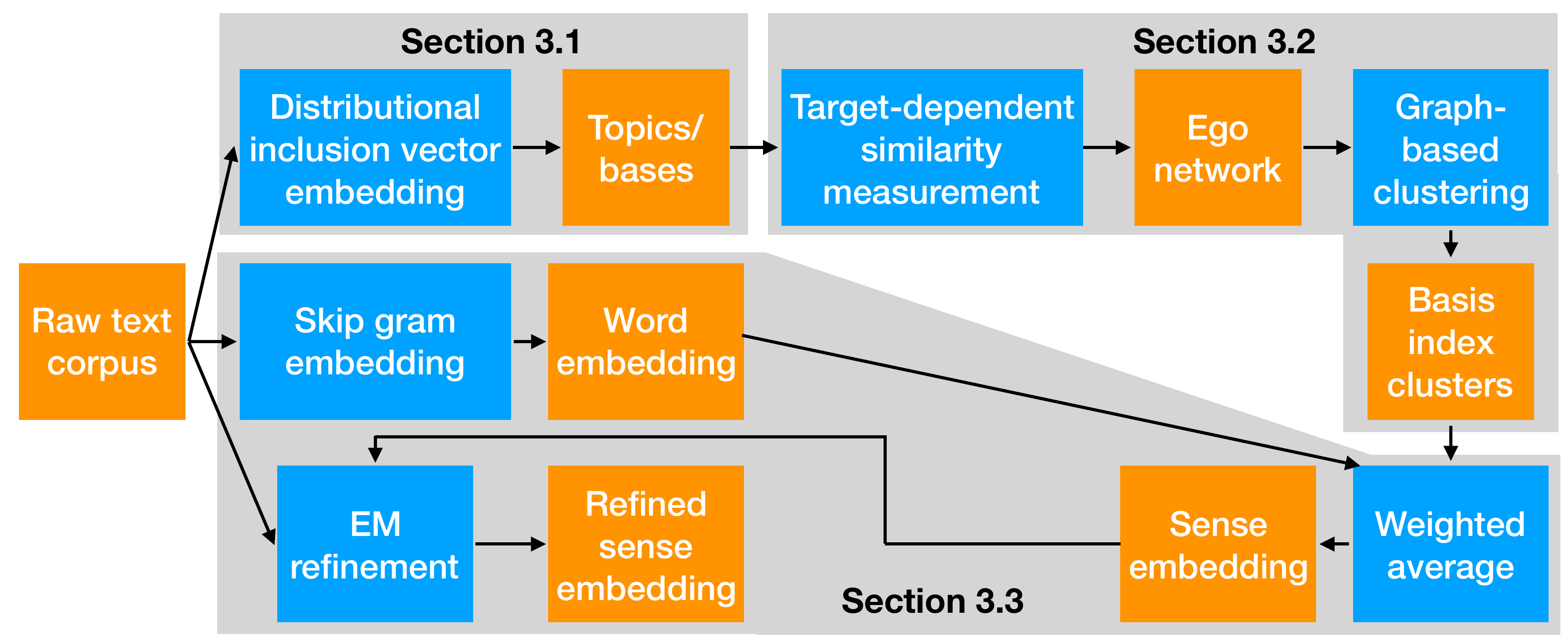}
\caption{The flowchart of the proposed method. The blue boxes are processing steps, the orange boxes are input and output data of each step, and the gray areas indicate the sections describing the included steps.}
\label{fig:WSI_flow_chart}
\end{figure*}

The flowchart of our method is illustrated in Figure~\ref{fig:WSI_flow_chart}. We will first briefly introduce distributional inclusion vector embedding (DIVE)~\citep{chang2017unsupervised} in Section~\ref{sec:DIVE}, illustrate how we use DIVE as a topic model to construct a graph in Section~\ref{sec:graph_clustering}, and after clustering the topics, we explain the way to converting each topic cluster to a sense embedding in Section~\ref{sec:cluster_2_emb}. 

\begin{figure*}[t!]
\centering
\includegraphics[width=0.95\linewidth]{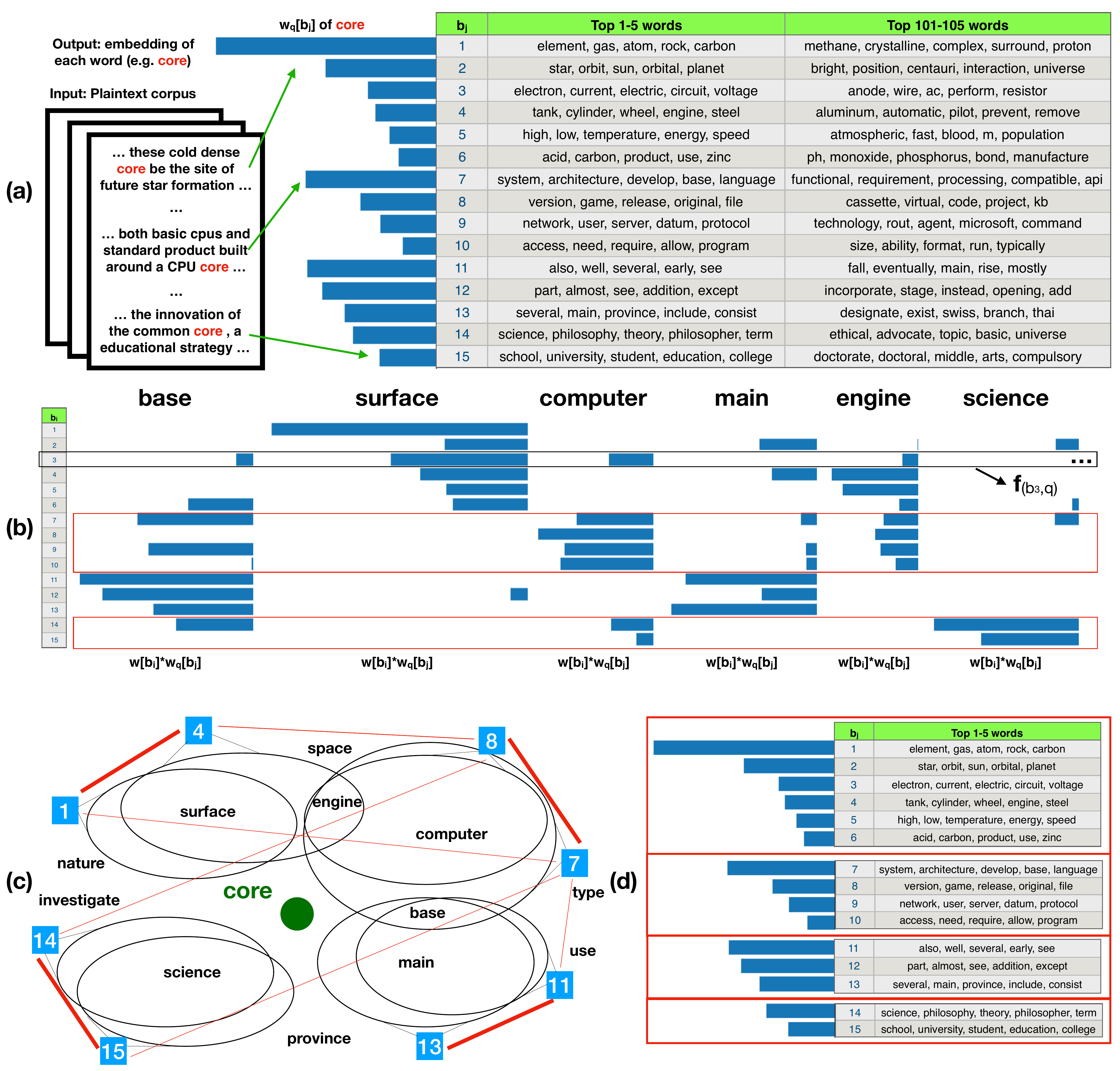}
\caption{A visualization of finding the senses of the word \smallt{core}. \\ \textbf{(a)} The DIVE $\mathbf{w}[b_j]$ of the word \smallt{core} (only top 15 basis indexes are shown). The words in each row of the table are sorted by its embedding value in the basis index. \\
\textbf{(b)} Weighted DIVE of six words on these 15 basis indexes relevant to \smallt{core} as the features for measuring the similarity between basis indexes. The two red boxes indicate two final clusters we discovered at the end within which the feature words tend to have similar embedding values. \\
\textbf{(c)} The ego network constructed for the word \smallt{core}. Each blue box and the corresponding circle represent a basis index or topic (only 8 out of 15 basis indexes are plotted and their index numbers $b_j$ are shown in the blue boxes), which is a node in the network. Two basis indexes are more similar if more relevant words (i.e., close to \smallt{core}) occur frequently in both corresponding topics. For example, the topic 7 and 8 are more similar because of the frequent appearance of the relevant words such as \smallt{computer} in both topics. The larger similarity is represented by a thicker red line. The ego network is a complete graph but only a subset of edges are plotted in the figure. \\
\textbf{(d)} The final clustering results when the number of clusters is set to be 4.}
\label{fig:embedding_vis_WSI}
\end{figure*}

\subsection{Distributional Inclusion Vector Embedding (DIVE)}
\label{sec:DIVE}

Distributional inclusion vector embedding (DIVE) is a variation of skip-gram model~\citep{mikolov2013distributed}. The two major differences compared with skip-gram are that 
\begin{enumerate*}[label=(\arabic*)]
  \item all word embeddings and context embeddings are constrained to be non-negative, and
  \item the weights of negative sampling for each word is inversely proportional to its frequency.
\end{enumerate*}
Specifically, the objective function of DIVE is defined by
\begin{equation}\label{eq-DIVE}
\small
\begin{aligned}
&l_{DIVE}=\sum_{w}\sum_{c}\#(w,c)\log\sigma(\mathbf{w}^T\mathbf{c}) ~+ \\
&k_I\sum_{w} \frac{Z}{\#(w)} \sum_{c}\#(w,c) \underset{c_N\sim P_D}{\mathbb{E}} [\log \sigma(-\mathbf{w}^T\mathbf{c_N})], 
\end{aligned}
\end{equation}
where the word embedding $\mathbf{w} \geq 0$, the context embedding $\mathbf{c} \geq 0, \mathbf{c_N} \geq 0$, $\#(w,c)$ are number of times context word $c$ co-occur with $w$, $\#(w) = \sum\limits_{c} \#(w,c)$, $\sigma$ is the logistic sigmoid function, $k_I$ is a constant hyper-parameter, $Z$ is the average $\#(w)$ of all words (i.e., $Z=\frac{\sum_w \#(w)}{|V|}$ and $|V|$ is the size of vocabulary), and $P_D$ is the distribution of negative samples. The two modifications do not change the time and space complexity of training skip-gram, which is one of the most scalable word embedding methods~\citep{levy2015improving}.

DIVE is originally designed to perform unsupervised hypernymy detection task, and its goal is to preserve the inclusion relation between two context features in the sparse bag of words (SBOW) representation. When the co-occurred context histogram of the word $y$ includes that of the word $x$, it means that for all context words $c$ in the vocabulary $V$, $c$ will co-occur more times with $y$ than with $x$. In this paper, the context words of a target word means the words co-occur with a target word mention within a small window in the corpus. The default context window size for DIVE is 10. \citet{chang2017unsupervised} show that the DIVE is able to compress the sparse bag of words while approximately preserving the inclusion in the low-dimensional space. Formally, 
\begin{equation}\label{eq-DIH}
\begin{aligned}
& \forall c \in V, \; \#(x,c) \leq \#(y,c)  \\
\tilde{\iff} & \forall i \in \{1,...,L \}, \; \mathbf{x}[i] \leq \mathbf{y}[i],
\end{aligned}
\end{equation}
where $\tilde{\iff}$ means approximately equivalent, $\#(x,c)$ and $\#(y,c)$ are number of times context word $c$ co-occurs with $x$ and $y$, respectively. $\mathbf{x}$ and $\mathbf{y}$ are the embeddings of the words $x$ and $y$, respectively, $\mathbf{x}[i]$ is the embedding value of in $i$th dimension (i.e., $i$th basis index). and $L$ is number of DIVE basis indexes. See \citet{chang2017unsupervised} for more the derivation of the equation.

In order to satisfy equation~\eqref{eq-DIH}, each basis index of DIVE corresponds to a topic and the embedding value at that index represents how often the word appears in the topic. This is because if the embedding of one word $y$ has higher value in every dimension (i.e., higher frequency in every topic) than the value of another word $x$, the context words $c$ in the topics usually co-occur more frequently with $y$ than with $x$. Inversely, if $x$ appears more often in one topic than $y$ (i.e., the embedding value of $x$ in the corresponding dimension is higher than that of $y$), some context words $c$ in the topic could co-occur more often with $x$ than with $y$.

%Let us assume that there are $L$ groups of context words in the corpus and all context words appear beside other context words in the same topic most of the time. A good basis vector that makes word embedding satisfy equation~\eqref{eq-DIH} is the frequency of a topic co-occurs with all words. 

In Figure~\ref{fig:embedding_vis_WSI} (a), we present three mentions of the word \smallt{core} and its top 15 basis indexes in DIVE. The word that has a higher value in a basis index is more frequent in the corresponding topic. For example, the top 1-5 words in the second column of the table look more frequent (and usually more general) than the top 101-105 words.

\subsection{Graph-Based Clustering}
\label{sec:graph_clustering}
For each target word, we build an ego network whose nodes are the basis indexes relevant to the word. The basis index $b$ is relevant if DIVE of the target word $q$ has a value $\mathbf{w}_q[b]$ higher than a threshold $T$. The threshold is set to be $1\%$ of average non-zero $\mathbf{w}_q[b]$ over basis indexes in our experiment. 

Every pair of nodes are linked by an edge weighted by the similarity between the two basis indexes. Each basis index $b_i$ is represented by a feature vector. A naive way to prepare the feature vector of $i$th basis index $\mathbf{f}_{(b_i)}$ is to use the embedding values in that index $\mathbf{w}[b_i]$ of all the words in our vocabulary $V$. That is,
\begin{equation}
\mathbf{f}_{(b_i)} = \underset{\mathbf{w} \in V}{\oplus} \mathbf{w}[b_i],
\end{equation}
the operator $\oplus$ means concatenation. However, as discussed in Section~\ref{sec:intro}, measuring similarity using the global features might group topics together based on the co-occurrence of words which are unrelated to the query words. Instead, we want to make the similarity dependent on the query word.

To create target-dependent similarity measurement, we only consider the embedding of words which are related to the query word as the features of basis indexes. Specifically, given a query word $q$, we only take the top $n$ words of every basis index $j$ in the set $B_j(n)$ instead of considering all the words in the vocabulary. Then, we weigh the feature based on how likely it is to observe the target word in topic $j$ ($\mathbf{w}_q[b_j]$) and concatenate all features together. That is, the feature vector of the $i$th dimension $\mathbf{f}_{(b_i,q)}$ is defined as:
\begin{equation}
\mathbf{f}_{(b_i,q)} = \overset{L}{\underset{j=1}{\oplus}} \underset{\mathbf{w} \in B_j(n)}{\oplus} \mathbf{w}[b_i] \cdot \mathbf{w}_q[b_j],
\end{equation}
where $n$ is fixed as 100 in the experiment. 
%$\mathbf{w}_q[b_j]$ is the value of $j$th basis index of the query word embedding, 

In addition to decreasing the weight of irrelevant words, we also lower the influence of irrelevant bases by defining the similarity between two basis indexes as
\begin{equation}
\begin{aligned}
SIM(b_i,b_j,q) = & \cos(\mathbf{f}_{(b_i,q)},\mathbf{f}_{(b_j,q)}) \cdot \\ &\log(\frac{\min(\mathbf{w}_q[b_i],\mathbf{w}_q[b_j])}{T}),
\end{aligned}
\end{equation}
where $\cos(\mathbf{f}_{(b_i,q)},\mathbf{f}_{(b_j,q)})$ is the cosine similarity between the features of two basis indexes, and the term $\log(\frac{\min(\mathbf{w}_q[b_i],\mathbf{w}_q[b_j])}{T})$ is to prevent irrelevant basis indexes in the ego network misleading the clustering algorithm. Notice that $SIM(b_i,b_j,q) \geq 0$ because all features $\mathbf{f}_{(b,q)} \geq 0$ and every node is a relevant basis index $b$ with $\mathbf{w}_q[b] > T$.

After the ego network is constructed, we could apply any hierarchical graph clustering. In this paper, we just choose spectral clustering with fixed number of clusters for simplicity. In our experiment, DIVE with 100 dimensions produces only 6.4 relevant basis indexes on average which needs to be clustered for each target word. This number goes to only 19 for DIVE with 300 dimensions. Thus, we are allowed to use spectral clustering to perform global optimization without inducing large computational overhead in this step.

In Figure~\ref{fig:embedding_vis_WSI}, we use the target word \smallt{core} as an example to illustrate our clustering algorithm. After DIVE is trained in (a), we visualize six dimensions of features for each basis index $\mathbf{f}_{(b_i,q)}$ in (b). Using the features, we can build the ego network as shown in (c). The figure highlights the novelty of our approach. Instead of directly clustering words as other graph-based methods, we group the words first and cluster the groups to form senses. Since the basis is global, we do not have to retrain it given a different target word. DIVE provides us an easy and efficient way to ignore the irrelevant words being far away from \smallt{core} in (c), such as \smallt{province} or \smallt{space}, and cluster based on the words close to the target word such as \smallt{main} or \smallt{computer}. The target-dependent similarity measurement preserves the main spirit of existing graph-based approaches.

\subsection{From Basis Index Clusters to Sense Embeddings}
\label{sec:cluster_2_emb}
As shown in Figure~\ref{fig:embedding_vis_WSI} (d), every sense is represented by a group of basis indexes each of which has a weight based on its relevancy to the target word (e.g., the relevancy of $b_i$th basis index is $\mathbf{w}_q[b_i]$). In order to apply existing WSI evaluation and potentially other downstream applications, we convert the basis index clusters to sense embedding.

First, we train a word embedding. Any existing embeddings could be used and we choose skip-gram due to its efficiency. Based on the trained word2vec, we first create a topic embedding for each basis index by averaging skip-gram embedding of the top 1000 words $B_i(1000)$ weighted by the DIVE $\mathbf{w}[b_i]$ of the words at $b_i$th basis index as given as:
\begin{equation}
    \mathbf{t}_{b_i} = \frac{\sum_{\mathbf{w} \in B_i(1000)} \exp(\mathbf{w'}[b_i]) \cdot \mathbf{e}_w }{\sum_{\mathbf{w} \in B_i(1000)} \exp(\mathbf{w'}[b_i])} \label{weight},
\end{equation}
where $\mathbf{e}_w$ is the skip-gram embedding for the word whose DIVE are $\mathbf{w}$, and $\mathbf{w'}$ is normalized DIVE such that its average $\frac{\sum_{\mathbf{w} \in B_i(1000)} \mathbf{w'}[b_i]}{1000}= 1$. We take exponential on $\mathbf{w'}[b_i]$ to focus on the words that are more important to the $b_i$ basis index because DIVE roughly models the $\log$ of word frequency in each topic~\citep{chang2017unsupervised}.

To generate $k$th sense embedding $\mathbf{s}^q_{k}$ for a target word $q$, we take the average of all the topic embeddings in the $k$th sense cluster (found in Section~\ref{sec:graph_clustering}) weighted by the relevancy between every topic and the target word. Specifically, 
\begin{equation}
    \mathbf{s}^q_{k} = \frac{\sum_{b_i \in S^q_k} \exp(\mathbf{w'}_q[b_i]) \cdot \mathbf{t}_{b_i}}{\sum_{b_i \in S^q_k} \exp(\mathbf{w'}_q[b_i])},
\label{sense}
\end{equation}
where $S^q_k$ is the set of basis indexes that belongs to the $k$th cluster, $\mathbf{w'}_q$ is normalized DIVE of the target word such that its average $\frac{\sum_{b_i \in N} \mathbf{w'}_q[b_i]}{|N|}= 1$, and $N$ is the set of nodes in the ego network. 

When converting clusters into embeddings, the previous graph-based WSI methods, such as ~\citet{PelevinaABP16}, average the embedding of related words. The average is effective in terms of discriminating the contexts of  target word mentions, but it might not be a good embedding for the sense of target word itself. For instance, one sense embedding of \smallt{core} could be close to the embedding of \smallt{computer}, but the \smallt{computer} embedding does not represent the sense of \smallt{core} in \smallt{computer} context as well as the embedding of \smallt{cpu}. Our method suffers the similar problem.

To solve the issue, we use the sense embeddings from clusters as the initialization of an expectation maximization (EM) refinement. At E-step, we predict the sense of every target token by checking which sense embedding the average word embedding of the current sentence is closest to, and assign the sense to the target token (e.g., \smallt{bank} $\rightarrow$ \smallt{bank\_1}). At M-step, we retrain the skip-gram using the updated corpus. Our refinement process could be seen as a simplified version of multi-sense skip-gram (MSSG)~\citep{NeelakantanSPM14}, which can be easily implemented using existing word embedding library.

\section{Experiments}
We first conduct a qualitative experiment to verify that our clustering algorithm performs well on some typical polysemy, and show the results in Table~\ref{tb:WSD}. As we can see, our method can not only separate two senses in very different contexts but also can distinguish more subtle sense difference such as identifying the \smallt{car} context and \smallt{competition} context as two different senses of the target word \smallt{race}.

Intuitively speaking, our method could be especially useful when it comes to increase the recall of less common senses (like discovering the educational meaning of \smallt{core}), but it is hard to verify the claim using existing WSI benchmarks because the common senses, especially the most frequent sense, often dominate in the benchmarks unless using the datasets where the bias is removed. In the following sections, we will first introduce the setup and then the experiments on 3 datasets.

\begin{table*}[t!]
\centering
%Spectral clustering on the non-zero DIVE dimensions of the query word. 
\scalebox{0.8}{
\begin{tabular}{|c|c|cc|}
\hline
Query& CID &  \multicolumn{2}{c|}{Top 5 words in the top dimensions} \\
\hline \hline
 \multirow{4}{*}{rock}  & \multirow{2}{*}{ 1 } & element, gas, atom, rock, carbon & sea, lake, river, area, water \\
& & find, specie, species, animal, bird & point, side, line, front, circle \\ \cline{2-4}
& \multirow{2}{*}{ 2 } & band, song, album, music, rock & write, john, guitar, band, author \\
& & early, work, century, late, begin & include, several, show, television, film \\ \hline \hline
\multirow{4}{*}{bank} & \multirow{2}{*}{1} & county, area, city, town, west & several, main, province, include, consist \\ %\cline{3-4}
 & & building, build, house, palace, site & sea, lake, river, area, water\\ \cline{2-4}
 & \multirow{2}{*}{2} & money, tax, price, pay, income & company, corporation, system, agency, service \\ %\cline{3-4}
 & & united, states, country, world, europe & state, palestinian, israel, right, palestine\\ \hline \hline
 \multirow{4}{*}{apple}  & \multirow{2}{*}{ 1 } & food, fruit, vegetable, meat, potato & goddess, zeus, god, hero, sauron \\
& & war, german, ii, germany, world & write, john, guitar, band, author \\ \cline{2-4}
& \multirow{2}{*}{ 2 } & version, game, release, original, file & car, company, sell, manufacturer, model \\
& & system, architecture, develop, base, language & include, several, show, television, film \\ \hline \hline
\multirow{4}{*}{star}  & \multirow{2}{*}{ 1 } & film, role, production, play, stage & character, series, game, novel, fantasy \\
& & wear, blue, color, instrument, red & write, john, guitar, band, author \\ \cline{2-4}
& \multirow{2}{*}{ 2 } & element, gas, atom, rock, carbon & star, orbit, sun, orbital, planet \\
& & give, term, vector, mass, momentum & light, image, lens, telescope, camera \\ \hline \hline
%\multirow{4}{*}{star} & \multirow{2}{*}{1} & film, role, production, play, stage & character, series, game, novel, fantasy \\ %\cline{3-4}
% & & wear, blue, color, instrument, red & write, john, guitar, band, author\\ \cline{2-4}
% & \multirow{2}{*}{2} & element, gas, atom, rock, carbon & sea, lake, river, area, water \\ %\cline{3-4}
% & & find, specie, species, animal, bird & point, side, line, front, circle\\ \hline \hline
%\multirow{4}{*}{cell}  & \multirow{2}{*}{ 1 } & may, cell, protein, gene, receptor & cause, disease, effect, infection, increase \\
%& & electron, current, electric, circuit, voltage & element, gas, atom, rock, carbon \\ \cline{2-4}
%& \multirow{2}{*}{ 2 } & form, vowel, word, name, call & network, user, server, datum, protocol \\
%& & tank, cylinder, wheel, engine, steel & access, need, require, allow, program \\ \hline \hline
%\multirow{4}{*}{left}  & \multirow{2}{*}{ 1 } & war, communist, lead, government, party & political, support, policy, issue, concern \\
%& & army, force, infantry, military, battle & state, republic, independent, finland, political \\ \cline{2-4}
%& \multirow{2}{*}{ 2 } & point, side, line, front, circle & head, leg, long, foot, hand \\
%& & take, give, place, player, hand & set, x, number, n, element \\ \hline \hline
\multirow{4}{*}{tank}  & \multirow{2}{*}{ 1 } & tank, cylinder, wheel, engine, steel & industry, export, industrial, economy, company \\
& & acid, carbon, product, use, zinc & network, user, server, datum, protocol \\ \cline{2-4}
& \multirow{2}{*}{ 2 } & army, force, infantry, military, battle & aircraft, navy, missile, ship, flight \\
& & however, attempt, result, despite, fail & war, german, ii, germany, world \\ \hline \hline
%\multirow{3}{*}{tank}  & 1  & tank, cylinder, wheel, engine, steel & aircraft, navy, missile, ship, flight \\ \cline{2-4}
%& 2  & acid, carbon, product, use, zinc & food, fruit, vegetable, meat, potato \\ \cline{2-4}
%& 3  & army, force, infantry, military, battle & however, attempt, result, despite, fail \\ \hline \hline
\multirow{3}{*}{race}  &  1 & win, world, cup, play, championship & two, one, three, four, another \\ \cline{2-4}
&  2 & railway, line, train, road, rail & car, company, sell, manufacturer, model \\ \cline{2-4}
&  3 & population, language, ethnic, native, people & female, age, woman, male, household \\ \hline \hline
\multirow{3}{*}{run}  &  1 & system, architecture, develop, base, language & access, need, require, allow, program \\ \cline{2-4}
&  2 & railway, line, train, road, rail & also, well, several, early, see \\ \cline{2-4}
&  3 & game, team, season, win, league & game, player, run, deal, baseball \\ \hline \hline
\multirow{3}{*}{tablet}  &  1 & bc, source, greek, ancient, date & book, publish, write, work, edition \\ \cline{2-4}
&  2 & use, system, design, term, method & version, game, release, original, file \\ \cline{2-4}
&  3 & system, blood, vessel, artery, intestine & patient, symptom, treatment, disorder, may \\ \hline 

%\multirow{3}{*}{fox}  &  1 & california, texas, american, states, united & letter, deliver, express, scandal, trial \\ \cline{2-4}
%&  2 & include, several, show, television, film & film, role, production, play, stage \\ \cline{2-4}
%&  3 & find, specie, species, animal, bird & head, leg, long, foot, hand \\ \hline \hline
%\multirow{3}{*}{core}  &  1 & system, architecture, develop, base, language & version, game, release, original, file \\ \cline{2-4}
%&  2 & also, well, several, early, see & part, almost, see, addition, except \\ \cline{2-4}
%&  3 & element, gas, atom, rock, carbon & star, orbit, sun, orbital, planet \\ \hline \hline

\end{tabular}
}
\caption{Examples of sense clusters on polysemous words. When the number of clusters is set to be 2, we present the top 4 basis indexes $b_j$ in each sense cluster, which have the highest values on the target word embedding $\mathbf{w}_q[b_j]$. Otherwise, the top 2 basis indexes are presented. CID refers to sense cluster ID. The top 5 words with the highest values of each basis index $\mathbf{w}[b_j]$ are presented.}
\label{tb:WSD}

\end{table*}

\subsection{Experimental Setup}
We train DIVE on first 51.2 million tokens of WaCkypedia~\citep{baroni2009wacky}, the dataset suggested by~\citet{chang2017unsupervised}, and the default hyper-parameter setting is used except the number of embedding dimensions $L$ (i.e., number of basis indexes). We train two DIVEs, one with 100 dimensions and the other with 300 dimensions to study how the granularity of basis affects the performance. For all other steps or baselines, we train them on the whole WaCkypedia where the stop words are removed.

%adopt the  and the training corpus suggested by to .
%(explain pre-trained DIVE embedding)

%wacky\citep{baroni2009wacky}
%number of DIVE dimension

For our clustering module, we use all the default hyper-parameters of the spectral clustering library in Scikit-learn 0.18.2~\citep{scikit-learn} except the number of clusters is fixed at 2. Setting a number larger than 2 makes it harder to compare with the results generated from other baselines whose default hyper-parameters usually make average number of senses between 1 and 2. During EM, we train the skip-gram embedding on the whole WaCkypedia where we treat every consecutive 20 tokens as a sentence, and the refinement stops after 3 EM iteration. In the tables of this section, our methods using DIVE with 100 and 300 dimensions are denoted as DIVE (100) and DIVE (300), respectively.
%number of senses

%Write training details: e.g., 
%EM: sentence with length 20
%how many EM iterations

In all quantitative experiments, we compare our method with \citet{PelevinaABP16}, a state-of-the-art graph-based clustering which builds ego graphs based on words similar to the target words, so we call it word graph (WG). To train the model, we first train skip-gram on whole WaCkypedia and use all the default hyper-parameters in their released code to get sense embeddings.\footnote{\url{https://github.com/tudarmstadt-lt/sensegram}} We also apply our EM refinement step to their output embedding to make the comparison fair and call this variation WG+EM. We also compare our method with the baseline which randomly assigns two senses to every token and performs EM to refine the embedding (i.e., only adopting our post-processing step). The method is similar to multiple-sense skip-gram~\citep{NeelakantanSPM14}, so we call it MSSG in our tables.

%During the testing time, each method is asked to 
In all datasets, evaluation involves the similarity measurement between a sense of the target word and a context. For each query, we compute cosine similarity between the context embedding and the sense embedding of the target word, where the context embedding $\mathbf{e}_c$ is the average embedding of word in the context. Notice that each word in the context could also be polysemous. In these cases, we adopt the sense embedding of the context word that is closest to the sense of the target word (i.e., highest cosine similarity). 

\subsection{Word Context Relevance (WCR)}
Given a target word, the task \citep{arora2016linear,sun2017simple} is to identify the true context corresponding to a sense of the target word out of 10 other randomly selected false contexts, where a context is presented by similar words. For example, two of the true contexts for the target word \smallt{bank} are \smallt{water,land,river,...} and \smallt{institution,deposits,money...}.
We use the R1 dataset from \citet{sun2017simple}, which consists of 137 word types and 535 queries. 

For each query pair (target word $w_q$, context $c$), we compute the similarity between each sense of target word $\mathbf{s}^q_k$ and the context $\mathbf{e}_c$, and choose the senses of the target word with maximal similarity (i.e., $SIM(w_q,\mathbf{e}_c) = \max_k \cos(\mathbf{s}^q_k,\mathbf{e}_c)$).  Then, we rank the similarity of 11 query pairs, which consist of 1 true context and 10 false contexts. The performance of different methods is evaluated by checking whether the top 1 (i.e., the pair with the highest similarity) is true. The metric~\citep{sun2017simple} is called Precision@1. 

The results are shown in Table~\ref{tb:R1}. Since the task is to identify the related contexts, skip-gram is a good baseline~\citep{sun2017simple}. In this dataset, each sense is equally important, regardless how often the sense appears in the corpus. The significantly better performance from DIVE demonstrates our capability of modeling more fine-grained senses of polysemous words.
%We found that our method performs significantly better than other baselines because our method

\begin{table}[t!]
\begin{center}
\begin{tabular} {|c|c|c|c|}
\hline
Skip-gram & WG & WG+EM   \\ \hline
%52.7 & 61.9 & 59.1  \\ \hline
52.7 & 42.1 & 59.1  \\ \hline
MSSG & DIVE (100) & DIVE (300) \\ \hline
60 & \textbf{63.2} & 62.6  \\ \hline
\end{tabular}
\caption{Precision@1 on the WCR R1 (\%).}
\label{tb:R1}
\end{center}
\end{table}

\begin{table}[t!]
\begin{center}
\scalebox{0.85}{
\begin{tabular} {|c|c|c|c|c|c|c|}
\hline
\multirow{2}{*}{Model}& \multicolumn{3}{|c|}{TWSI} & \multicolumn{3}{|c|}{balanced TWSI}\\ \cline{2-7}

 & P & R & F1 & P & R & F1 \\ \hline
MSSG rnd & 66.1&65.7&65.9 & 33.9&33.7&33.8 \\ \hline
MSSG & 66.2 & 65.8 & 66.0 & 34.3 & 34.2 & 34.2 \\ \hline
WG & \textbf{68.6} & \textbf{68.1}  & \textbf{68.4} & 38.7 & 38.5 & 38.6 \\ \hline
WG+EM & 68.3 & 67.8 & 68.0 & 38.4 & 38.2 & 38.3 \\ \hline
DIVE rnd & 63.4&63.0&63.2 & 33.4&33.2&33.3\\ \hline
DIVE (100) & 67.6 & 67.2 & 67.4 & \textbf{39.7} & \textbf{39.5} & \textbf{39.6}\\ \hline
DIVE (300) & 67.4 & 66.9 & 67.2 & 39.0 & 38.8 & 38.9 \\
\hline
\end{tabular}
}
\caption{Results obtained on the TWSI task (\%), where P is precision and R is recall. MSSG rnd and DIVE rnd are baselines which randomly assign sense given inventory built by MSSG and DIVE, respectively.}
\label{tb:TWSI}
\end{center}
\end{table}

\begin{table}[t!]
\begin{center}
\scalebox{0.85}{
\begin{tabular} {|c|c|c|c|c|c|}
\hline
Model & JI & Tau & WNDCG & FNMI & FB-C  \\ \hline
All-1 &19.2&60.9&28.8&0&\textbf{62.3} \\ \hline
Rnd & 21.8&62.8&28.7&2.8&47.4 \\ \hline
MSSG & \textbf{22.2}&\textbf{62.9}&29.0&3.2&48.9  \\ \hline
WG & 21.2&61.2&29.0&1.6&58.1  \\ \hline
WG+EM & 21.0&61.5&29.0&1.3&57.8  \\ \hline
DIVE (100) & 21.9&61.9&\textbf{29.3}&3.1&50.6 \\ \hline
DIVE (300) & 22.1&62.8&29.1&\textbf{3.5}&49.9  \\
\hline
\end{tabular}
}
\caption{Results obtained on the SemEval 2013 task (\%), where JI is Jaccard Index, FNMI is Fuzzy NMI, and FB-C is Fuzzy B-Cubed. All-1 is to assign all senses to be the same and Rnd is to randomly assign all senses to 2 groups.}
\label{tb:SemEval}
\end{center}
\end{table}

%TWSI\citep{Biemann12}
\subsection{TWSI Evaluation}
The Turk bootstrap Word Sense Inventory (TWSI) task \citep{Biemann12} is based on a large dataset, which consists of 1,012 nouns accompanied with 145,140 context sentences. The task is to identify the correct sense of the target nouns, and all WSI algorithms choose the sense whose embedding is most similar to the context embedding. 
%Each sense in the TWSI task is represented by a bag-of-words. 

Dataset is heavily skewed with 79\% of contexts being assigned to the most frequent senses. To remove this bias, we follow the procedure described in \citet{PelevinaABP16} to create balanced TWSI. Specifically, we only keep the first 5 contexts of each sense of every target word to make every sense count equally. The procedure yields 8710 pairs of senses and contexts. 
%We remove the sense which does not have 5 contexts and remove words which only have 1 sense after the pre-processing. 
%is assigned equal number of contexts
%only 5 context 

%we perform the same evaluations on a balanced subset of the dataset. To allow comparisons, we use the same subset to evaluate as .
When evaluating on TWSI, each method needs to represent the sense by a sparse bag-of-word context feature called sense inventory. The evaluation script\footnote{\url{https://github.com/tudarmstadt-lt/context-eval}} first maps each sense predicted by each algorithm to a ground truth sense. Then, the problem becomes a classification task, which can be evaluated by precision, recall, and F1.
%we will create an sense inventory b

In Table~\ref{tb:TWSI}, we can see that DIVE performs slightly worse than WG~\citep{PelevinaABP16} in full TWSI, but becomes slightly better in balanced TWSI. We suspect this is because our number of sense is 2 but the WG generates the output where the average number of senses is around 1.5, which might do better when a sense of each word occurs most of the time. Notice that the comparison in balanced TWSI is fair because the experiments in~\citet{PelevinaABP16} show that WG performs worse when increasing number of clusters. The results also suggest that a sufficient number of basis vectors seldom group two senses together (otherwise, increasing the resolution/dimension of DIVE should be helpful).

%more dimension does not get better. Showing that small number of dimension could be enough and efficient

%To perform the word sense disambiguation we follow the following process. We are given a target noun and its context sentence. Each context-word itself has multiple possible embeddings based on multiple senses of the context-word. The first challenge is to encode the context into the correct embeddings. To do this we generate all combinations of the context encodings and compare them to the each of the possible sense embeddings of the target noun by taking their cosine similarity. The context embedding is generated by taking the average over the most similar combination. Once we have the context embedding we simply take the cosine similarity of this embedding with all possible sense embeddings of the target noun. 
%The most similar sense is said to be the sense being used. We then map the predicted sense to the TWSI sense inventory and calculate the precision, recall and F-scores. 

\subsection{SemEval-2013 task 13 Evaluation}
%Semeval-2013 task 13\citep{jurgens2013semeval}
SemEval-2013 task 13~\citep{jurgens2013semeval} provides a smaller dataset which consists of 50 words which include nouns, verbs, and adjectives. The context prediction is done in the same way as TWSI, and the meaning of each metric could be found in~\citet{jurgens2013semeval}. In Table~\ref{tb:SemEval}, we can see our method performs roughly the same compared with other baselines.

\section{Conclusions}
We propose a novel graph-based WSI approach. In order to save the time of performing a nearest neighbor search, we first group words into basis/topics using distributional inclusion vector embedding (DIVE), compute target-dependent similarity between basis indexes, and then perform graph clustering. Our experimental results show that the method achieves the state-of-the-art performances and is able to capture less common senses with higher accuracy.

\section{Acknowledgement}
This material is based upon work supported in part by the Center for Data Science and the Center for Intelligent Information Retrieval, in part by Lexalytics, in part by the Chan Zuckerberg Initiative under the project Scientific Knowledge Base Construction, and in part by the National Science Foundation under Grant No. IIS-1514053 and No. DMR-1534431, in part using high performance computing equipment obtained under a grant from the Collaborative R\&D Fund managed by the Massachusetts Technology Collaborative. Any opinions, findings and conclusions or recommendations expressed in this material are those of the authors and do not necessarily reflect those of the sponsor.

%\bibliography{naaclhlt2018}
\bibliography{ref}
\bibliographystyle{acl_natbib}

%\appendix

%\newpage
%\clearpage

%\section{Supplemental Material}
%\label{sec:supplemental}
%\input{appendix}

\end{document}